# Neural Oscillations for Encoding and Decoding Declarative Memory using EEG Signals

Jenifer Kalafatovich[1], Minji Lee[1]
[1]Department of Brain and Cognitive Engineering, Korea University, Seoul, Korea
jenifer@korea.ac.kr, minjilee@korea.ac.kr

*Abstract*—**Declarative memory has been studied for its relationship with remembering daily life experiences. Previous studies reported changes in power spectra during encoding phase related to behavioral performance, however decoding phase still needs to be explored. This study investigates neural oscillations changes related to memory process. Participants were asked to perform a memory task for encoding and decoding phase while EEG signals were recorded. Results showed that for encoding phase, there was a significant decrease of power in low beta, high beta bands over fronto-central area and a decrease in low beta, high beta and gamma bands over left temporal area related to successful subsequent memory effects. For decoding phase, only significant decreases of alpha power were observed over fronto-central area. This finding showed relevance of beta and alpha band for encoding and decoding phase of a memory task respectively.**

## I. Introduction

Memory is a cognitive process that enables us to store information which can be recalled later as required, however, many diseases can alter memory abilities (dementia, mild cognitive disease, etc.). Therefore, understanding memory process and related brain mechanisms can be helpful to treat memory dysfunction disorders and improve learning efficiency [1]. Declarative memory refers to the ability to recall past events, facts and general knowledge [2], usually, declarative memory tests include a delayed retrieval. Delayed retrieval (30 min approximately between study and retrieval task during which participants have to perform a distraction task) in a memory task highly influences memory formation within long-term storage in healthy patients [3]. Thus, the study of declarative memory task with delayed retrieval can give us a better understanding of the mechanism involve remembering important information for daily life and its possible long-term storage.

Functional magnetic resonance imaging (fMRI) machine and electroencephalogram (EEG) have been widely used in attempts to understand cognitive tasks [4,5], and consciousness [6-8], the diagnosis of diseases [9-11], development of brain-computer interfaces, etc. [12-17], however the use of fMRI can be somehow inconvenient to use in daily life applications due to its size and its cost. On the other hand, EEG is a portable and inexpensive device to acquire brain signals, which make it very suitable for use in real-life [18-20]. Previous studies related to understand the memory process also involve the use of these devices [21-23]. Studies that involve the use of EEG either performed a statistical frequency band during only encoding phase (presentation of analysis comparing characteristic signals present in different facilitating words to be remembered later) [23, 24]; or attempted to predict subsequent memory effect reporting a low classification accuracy [25]. Additionally, the given task on most of these studies often integrates a declarative memory task with an associative memory task (that creates a relationship between two different stimuli, sometimes facilitating retrieval) consequently reported results to integrate EEG activity during both tasks. Therefore, the EEG activity during only declarative memory tasks is unclear yet.

Brain waves, in the field of memory, had been used in behavioral analysis, providing an analysis of frequency bands characteristics when the shown stimulus is later remembered or forgotten. Many studies related to declarative memory tasks reported significant changes in low beta and alpha band over frontal middle and parietal regions [23, 26] and power increase in gamma band of EEG signals during encoding phase [27], without performing the analysis of EEG signal during decoding phase. Therefore, it is possible to use this information to predict whether a presented stimulus will be successfully encoded. If a certain stimulus is delivered when a likely to remember brain state is present, then performance in memory task can be improved [28]. This opens the possibility of memory abilities enhancement, which was aimed before on studies that use transcranial direct current stimulation (tDCS), transcranial magnetic stimulation (TMS) or neurofeedback as methods to change participant's cortical excitability of specific brain regions [29-31].

This study explores the differences of neural oscillations not only during encoding but also decoding phase (recall test of previously presented words) of a declarative memory task. Encoding phase has been widely explored; however decoding phase needs to be studied in order to clarify: the brain state when a successful retrieval of shown stimulus takes place and the frequency band characteristics related to it. We compared EEG signals of successful remembered and forgotten words during a pure declarative memory task and analyze differences in neural oscillations over different brain regions in order to disentangle brain activity during a single type of memory task, specifically, a declarative memory task. These findings could be the first step to improve memory abilities.

This work was supported by Institute for Information & Communications Technology Promotion (IITP) grant funded by the Korea government (No. 2017-0-00451, Development of BCI based Brain and Cognitive Computing Technology for Recognizing User's Intentions using Deep Learning).

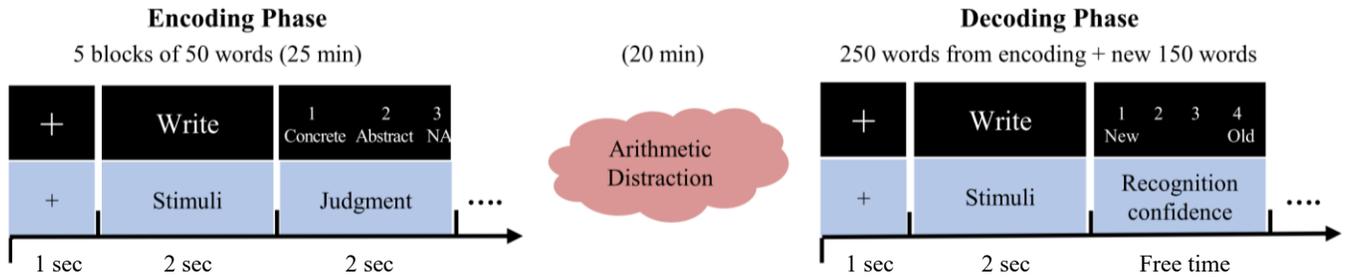

Figure 1. Experimental paradigm design. The declarative memory task is composed of an encoding and decoding phase, and an arithmetic distraction task.

## II. METHODS

### A. Participants

Seven right-handed healthy participants (3 females, 21-31 years old) with normal or correct-to-normal vision and no history of neurological disease were recruited for the study. Recruitment was done among students of Korea University, everyone had previous experience of English education. All the experiments were conducted according to the principles described in the Declaration of Helsinki. This study was reviewed and approved by the Institutional Review Board at Korea University (IRB-2019-0215).

### B. Experimental Setup

The experimental paradigm consists of two phases: encoding and decoding phases, separated by an arithmetic distraction task to avoid rehearsal of presented words [24]. Specifically, participants were asked to count backward from 1000 to zero in steps of seven for 20 min.

Figure 1 shows a schematic of the experimental paradigm. This paradigm was implemented using Psychophysics Toolbox. During the encoding phase, participants were shown five lists of fifty English nouns each one, randomly selected from a pool of 3,000 most commonly used nouns according to Oxford University (https://www.oxfordlearnersdictionaries.com). Each word was presented during 2 sec, preceded by a fixation cross that lasted 1 sec. The participant was asked to select whether the presented word was an abstract or a concrete noun using the keyboard, independently of the answer (abstract, concrete, or no answer at all) next trial began after 2 sec. Each list was separated by a blank screen of 5 sec.

For the decoding phase, all previous words presented during the encoding phase were presented again along with 150 new words in random order. A fixation cross was presented during 1 sec followed by the presentation of stimulus during 2 sec, participants had to decide whether the presented word was old or new using a confidence scale ranging from 1 (certain new) to 4 (certain old). Trials were separated by a black screen for 1 sec. Since free time was given for the recognition phase, the total duration of decoding phase varied from subject to subject having as average 27 min.

### C. Data Acquisition

EEG signals were recorded from 62 electrodes attached to the scalp according to the international 10–20 system using BRAINAMP (Brain Products, Germany) at a sampling rate of 1,000 Hz with the ground on Fpz and reference placed on FCz. All channels impedances were set below 10kΩ prior to signal measurement.

### D. Data Analysis

For EEG data analysis trials of encoding and decoding phases were separated into two conditions, successful remembered and forgotten trials. EEG signals were band-pass filtered from 0.5 to 40 Hz using a fifth-order Butterworth filter and down-sampled to 250 Hz. Trials were epoched concerning stimulus onset in intervals of 1 sec prior stimulus and 2 sec on-going stimulus and baseline corrected over the whole period.

Grand average over trials was calculated and compared event-related potentials (ERP) in the two conditions. Time frequency analysis was performed using fast Fourier transform on time bins of 100 msec from -1000 msec pre-stimulus to 1000 msec on-going stimulus. EEG power spectra were computed and averaged into five frequency bands: theta (3-7 Hz), alpha (7-13 Hz), low beta (13-17 Hz), high beta (17-30 Hz), gamma (30-40 Hz) bands.

### E. Statistical Analysis

We performed the Wilcoxon rank-sum non-parametric tests to investigate the differences in memory performance and reaction times between remembered and forgotten trials during encoding and decoding phase. In addition, Kruskal-Wallis test was applied to compare the averaged power spectra over pre-frontal, fronto-central, parietal, left temporal and right temporal areas across conditions (remembered and forgotten words during pre-stimulus and on-going stimulus). For post-hoc tests, Wilcoxon rank-sum tests were performed. All significance level is 0.05 with Bonferroni correction.

## III. RESULTS

### A. Beharioral Results

Table I shows the behavioral performance per participant. They were able to successful remember on average 84.8 ± 6.54% of presented words with a false positive rate of 18.6 ± 5.89%. During encoding phase 78.25 ± 18% of words were classified either as concrete or abstract words.

Reaction time was measured during encoding and decoding phases. For encoding phase, reaction time (time after the stimulus disappeared and participants select nature of presented word) for later successful remembered words were $780 \pm 18$ msec and for later forgotten words $820 \pm 19$ msec ($p = 0.07$). There was no relationship between encoding reaction time and later remember words. Figure 2 shows the averaged reaction time during decoding phase. For decoding phase, reaction time (time participants take to select confidence levels) for successful remembered words was $510 \pm 30$ msec, forgotten words $1030 \pm 69$ msec and false remembered words 720 msec. There was a significant difference between reaction time for remembered and forgotten words ($p = 0.03$). Participants took more time to select a confidence range for forgotten words.

TABLE I. MEMORY TASK PERFORMANCE DURING DECODING PHASE

| Subjects | Remember | Forgotten | False remember |
|---|---|---|---|
| Sub01 | 92.8% | 7.2% | 13.3% |
| Sub02 | 72.0% | 28.0% | 24.7% |
| Sub03 | 88.4% | 11.6% | 17.3% |
| Sub04 | 88.4% | 11.6% | 20.7% |
| Sub05 | 85.6% | 14.4% | 28.7% |
| Sub06 | 78.8% | 21.2% | 13.3% |
| Sub07 | 87.6% | 12.4% | 12.0% |
| *Mean* | *84.8%* | *15.2%* | *18.6%* |

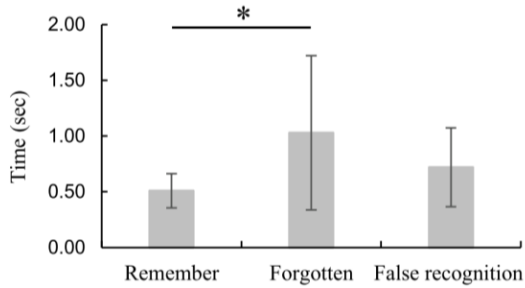

Figure 2. The averaged reaction time during decoding phase.* means *p*-value < 0.05.

### B. EEG Results

Figure 3 shows the scalp evolution of ERP difference between remembered and forgotten words during decoding phase. There were significant differences between pre-stimulus and on-going stimulus period were found over frontal electrodes.

EEG data were compared between remembered and forgotten trials over all subjects. Table II and Table III show chi-square and *p*-value of four conditions (activity during remember and forgotten trials divided on pre and on-going stimulus) over five brain regions and frequency bands during encoding and decoding phases, respectively.

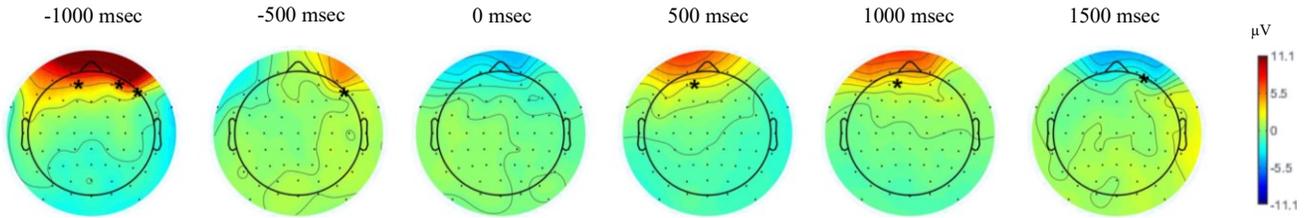

Figure 3. Topography of difference in amplitude between remembered and forgotten words during decoding phase. * means significant differences between two words.

TABLE II. *P*-VALUE OF CONDITIONS COMPARISON ACROSS BRAIN AREAS OF BAND POWER DURING ENCODING PHASE

| Brain regions / Frequency | Pre-frontal | | Fronto-central | | Parietal | | Left temporal | | Right temporal | |
|---|---|---|---|---|---|---|---|---|---|---|
| | chi-sq | *p*-value | chi-sq | *p*-value | chi-sq | *p*-value | chi-sq | *p*-value | chi-sq | *p*-value |
| Theta | 0.48 | 0.923 | 6.23 | 0.100 | 1.58 | 0.660 | 7.81 | 0.060 | 0.29 | 0.913 |
| Alpha | 0.55 | 0.907 | 7.84 | 0.049 | 2.15 | 0.541 | 7.85 | 0.054 | 0.24 | 0.971 |
| Low beta | 0.23 | 0.972 | 15.62 | 0.001 | 5.18 | 0.159 | 15.21 | 0.001 | 0.30 | 0.961 |
| High beta | 0.24 | 0.970 | 13.91 | 0.003 | 5.80 | 0.122 | 15.86 | 0.001 | 0.49 | 0.921 |
| Gamma | 0.70 | 0.872 | 9.81 | 0.020 | 6.35 | 0.095 | 16.86 | 0.008 | 0.36 | 0.948 |

TABLE III. *P*-VALUE OF CONDITIONS COMPARISON ACROSS BRAIN AREAS OF BAND POWER DURING DECODING PHASE

| Brain regions / Frequency | Pre-frontal | | Fronto-central | | Parietal | | Left temporal | | Right temporal | |
|---|---|---|---|---|---|---|---|---|---|---|
| | chi-sq | *p*-value | chi-sq | *p*-value | chi-sq | *p*-value | chi-sq | *p*-value | chi-sq | *p*-value |
| Theta | 4.60 | 0.203 | 10.23 | 0.016 | 3.91 | 0.271 | 1.05 | 0.789 | 3.16 | 0.367 |
| Alpha | 4.28 | 0.233 | 12.04 | 0.007 | 4.11 | 0.249 | 1.05 | 0.789 | 3.09 | 0.377 |
| Low beta | 1.94 | 0.585 | 11.64 | 0.008 | 4.27 | 0.234 | 1.35 | 0.716 | 3.41 | 0.333 |
| High beta | 1.73 | 0.629 | 9.81 | 0.020 | 5.89 | 0.117 | 6.56 | 0.087 | 3.35 | 0.340 |
| Gamma | 3.01 | 0.390 | 9.38 | 0.024 | 7.56 | 0.049 | 6.95 | 0.073 | 3.68 | 0.297 |

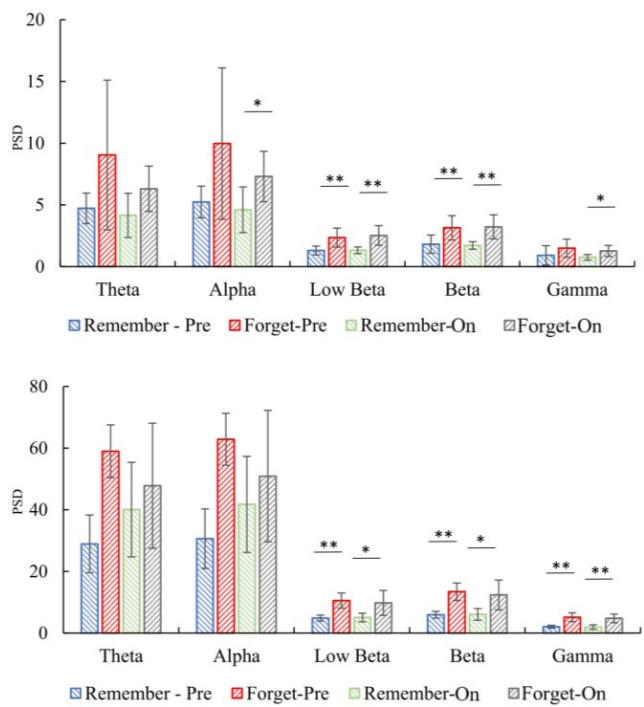
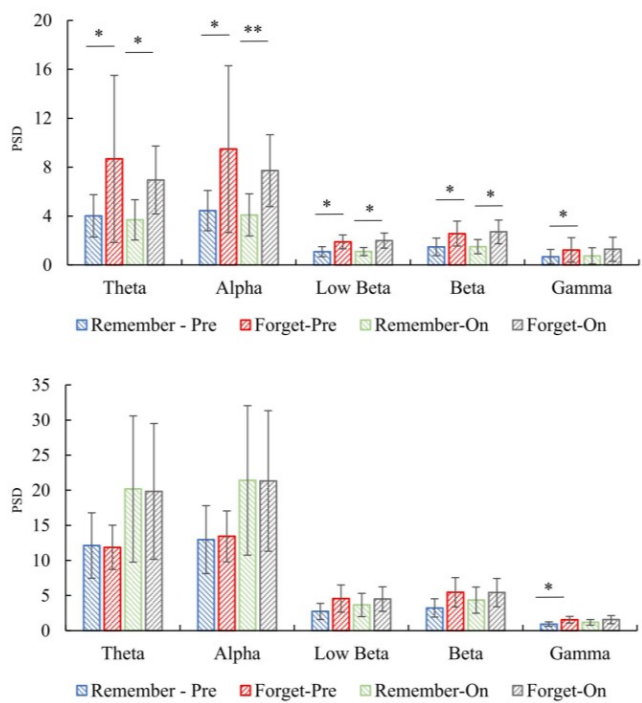

Figure 4. Difference between conditions over (up) fronto-central and (down) left temporal regions during encoding phase. * indicates $p < 0.05$ with no correction and ** indicates $p < 0.05$ with Bonferroni correction. PSD = power spectral density.

Figure 5. Difference between conditions over (up) fronto-central and (down) parietal regions during decoding phase. * indicates $p < 0.05$ with no correction and **indicates $p < 0.05$ with Bonferroni correction. PSD = power spectral density.

Figure 4 shows band power average values of different conditions in fronto-central and left temporal regions during encoding phase. For encoding phase, during pre-stimulus period a significant decreased of low beta and high beta power in fronto-central electrodes and a decreased of low beta, high beta and gamma bands in left temporal electrodes was found. For on-going stimulus period, statistical analysis revealed a significant decreased of power over alpha, low beta, high beta, and gamma bands in fronto-central electrodes, and a decreased of low beta, high beta and gamma bands in left temporal electrodes.

Figure 5 shows band power average values of different conditions in fronto-central and parietal regions during decoding phase. For decoding phase, during pre-stimulus period a significant decreased of theta, alpha, low beta, high beta and gamma power in fronto-central electrodes and a decreased of gamma power in parietal electrodes was found. For on-going stimulus period, there was a significant decreased of power over theta, alpha, low beta, and high beta bands in fronto-central electrodes.

## IV. CONCLUSION AND DISCUSSION

The present study examines the relationship between successful remembered and forgotten words during a declarative memory task. Difference in neural oscillations was characterized by a significant decreased during pre-stimulus and on-going period of low-beta and high beta power over fronto-central area; a decrease in gamma band during pre and on-going stimulus; and a decrease of low beta, and high beta powers during pre-stimulus over left temporal area. These results are in line with previous studies [24-26, 32], changes in low beta band over fronto-central electrodes could be related of the expectancy of a stimulus as showed in [20], however low beta band have also been related to the ability to form new memory [24] as it is related to attentional process [33]. In other works, temporal activity has even associated with the memory consolidation process [3].

For decoding phase, successful recognition was characterized by a significant decreased during on-going stimulus on alpha power over fronto-central electrodes, which can be related to attention mechanism [27] present during a successful retrieval. In other studies, alpha band decrease was related to semantic processing [26]. Consequently, decrease in alpha band during decoding phase can be attributed to a memory process and attentional increase.

One limitation of this study is the high difference in the amount of trials between successful remember and forgotten words. Normally, the average in performance on this kind of test is around 70% for successful remember stimuli [23, 24]. In our case, participants were able to remember 84.8% of presented words as average, this could be explained by the nature of presented stimuli (most commonly used words which can make them easy to remember). Additionally, to ensure long-term storage of words it will be necessary to perform a re-test of given words after a long period, however, this implicates repetitive visits and adequate time for evaluation, that is why it is rarely used due to the high time consuming [3].

In conclusion, this study differentiates neural oscillations during memory tasks, which arise the possibility that depending on brain states, the presented stimulus is more likely to be remembered or forgotten. Previous studies that modulate neural oscillations by stimulations (tDCS or TMS) or neurofeedback have shown some improvement in memory performance however results vary from study to study. We decide to investigate further the relationship between neural oscillations during encoding and decoding process. Also disentangled neural oscillations for a successful encoding and a successful retrieval and analyze whether performance changes depend on brain state that participants are in when the stimulus is shown and when retrieval test takes place.